\documentclass{article}
\pdfoutput=1

\usepackage{arxiv}

\usepackage[utf8]{inputenc} 
\usepackage[T1]{fontenc}    
\usepackage{hyperref}       
\usepackage{url}            
\usepackage{booktabs}       
\usepackage{amsfonts}       
\usepackage{nicefrac}       
\usepackage{microtype}      
\usepackage{graphicx}
\usepackage{subfig}
\usepackage{doi}
\usepackage{multirow}
\usepackage{multicol}
\usepackage{xcolor}
\usepackage{threeparttable}

\title{ETDPC: A Multimodality Framework for Classifying Pages in Electronic Theses and Dissertations}

\author{Muntabir Hasan Choudhury\\
	Old Dominion University\\
	Norfolk, VA 23529 \\
	\texttt{mchou001@odu.edu} \\
    \And
    Lamia Salsabil\\
    Old Dominion University\\
    Norfolk, VA 23529\\
    \texttt{lsals002@odu.edu} \\
	\And
	William A. Ingram \\
	Virginia Tech \\
	Blacksburg, VA 24061 \\
	\texttt{waingram@vt.edu} \\
	\And
	Edward A. Fox \\
	Virginia Tech \\
	Blacksburg, VA 24061 \\
	\texttt{fox@vt.edu} \\
    \And
    Jian Wu\\
	Old Dominion University\\
	Norfolk, VA 23529 \\
	\texttt{j1wu@odu.edu} \\
}

\renewcommand{\shorttitle}{ETDPC: A Multimodality Framework for Classifying Pages in ETDs}

\begin{document}
\maketitle

\begin{abstract}
Electronic theses and dissertations (ETDs) have been proposed, advocated, and generated for more than 25 years. Although ETDs are hosted by commercial or institutional digital library repositories, they are still an understudied type of scholarly big data, partially because they are usually longer than conference proceedings and journals. Segmenting ETDs will allow researchers to study sectional content. Readers can navigate to particular pages of interest, discover, and explore the content buried in these long documents. Most existing frameworks on document page classification are designed for classifying general documents and perform poorly on ETDs. In this paper, we propose ETDPC. Its backbone is a two-stream multimodal model with a cross-attention network to classify ETD pages into 13 categories. To overcome the challenge of imbalanced labeled samples, we augmented data for minority categories and employed a hierarchical classifier. ETDPC outperforms the state-of-the-art models in all categories, achieving an F1 of 0.84 -- 0.96 for 9 out of 13 categories.
We also demonstrated its data efficiency. The code and data can be found on GitHub\footnote{\url{https://github.com/lamps-lab/ETDMiner/tree/master/etd_segmentation}}.
\end{abstract}


\section{Introduction}
Electronic theses and dissertations (ETDs) are scholarly works of students pursuing higher education and successfully meeting the partial requirement of academic degrees. Since 1997, pioneered by Virginia Tech, many universities started requiring degree candidates to submit their ETDs, hosted by the university libraries or a centralized system such as ProQuest (recently acquired by Clarivate). ETDs have distinct features compared with conference papers and journal articles. They are book-length documents (i.e., typically 100 -- 400 pages long), and the topics may shift across chapters. In addition, ETDs have unique metadata fields (e.g., advisor, discipline, department) compared with regular scholarly papers. However, most ETD repositories still have limited tools and services for discovering and accessing the content and knowledge in ETDs. One step toward better content and knowledge discovery is to segment the entire ETD by content types so
information 
can be further mined using a customized content reader.

\begin{figure}[ht]
    \centering
    \includegraphics[width=0.774\textwidth]{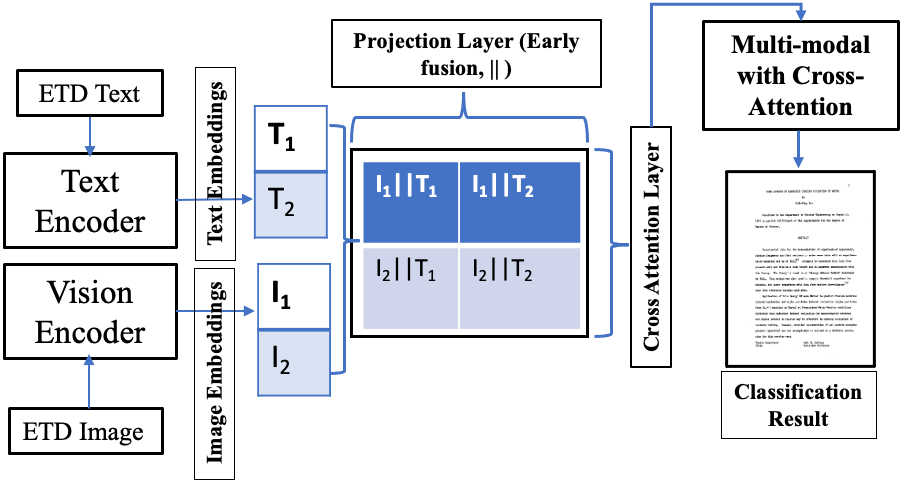}
    \caption{ETDPC -- A Multimodality Framework (I -- Image, and T -- Text).} 
    \label{fig:framework}
\end{figure}

ETDs can be scanned or born-digital (e.g., LaTeX), with complex document structures.
There are varied resolutions of scanned images, from typewritten and handwritten texts containing noise.
There are few
training samples available for classification tasks.
To address these challenges, we previously (\cite{ahuja-etal-2022-parsing}) contributed datasets and methods to segment ETDs using a bottom-up strategy. The method automatically annotated major structural components but still does not perform well in detecting minority classes (e.g., date, degree, equation, algorithms) due to a lack of training samples. 
Hence we considered multi-modality state-of-the-art (SOTA) frameworks (\cite{xu-etal-2021-layoutlmv2,Appalaraju_2021_ICCV}) which were fine-tuned on downstream tasks such as document image classification. These frameworks had been evaluated on the RVL-CDIP dataset (\cite{rvl-cdip}), consisting of scanned document images belonging to 16 classes (e.g., letter, form, email). However, even fine-tuned on ETDs, they do not generalize well on ETDs (e.g., achieving 9\% accuracy), and retraining them is non-trivial because of the lack of data. Moreover, text-based classification ignores the information encoded in the layout.
Therefore, we take a top-down approach by designing a new framework called ETD Page Classifier (ETDPC -- see Figure \ref{fig:framework}),
and apply it to
ETD pages, relative to 13 categories (e.g., title-page, chapters, dedication), using multimodality with a cross-attention network. Our contributions are below.
\begin{itemize}
    \item We proposed a two-stream multimodal classification model with cross-attention that uses a vision encoder (e.g., ResNet-50v2) and a text encoder (e.g., BERT with Talking-Heads Attention);
    \item We proposed a method to augment minority ETD pages leveraging paraphrasing techniques and image-based transformation;
    \item We built the ETD500 dataset with 92,371-page annotations of ETDs, plus PNGs, text, and bounding boxes; 
    \item We quantitatively demonstrated our system's robustness in classifying ETDs using both original and pseudo-training samples.
\end{itemize}

\section{Related Work}
In layout analysis for general documents, the existing frameworks usually adopt bottom-up approaches to identify document formats (forms, receipts, etc.). Several frameworks, including LayoutLM \cite{layoutlm}, LayoutLMv2 \cite{xu-etal-2021-layoutlmv2}, and DocFormer \cite{Appalaraju_2021_ICCV} used multimodality, introduced several pre-training tasks to the model, and then fine-tuned for downstream tasks.

These frameworks differ in their methodological approach and pre-training tasks. The authors in both LayoutLM \cite{layoutlm} and LayoutLMv2 \cite{xu-etal-2021-layoutlmv2} used the joint multimodal, where vision and text features are concatenated into one long sequence and then fed through a transformer self-attention layer to learn the cross-modality interaction between visual and textual information. 
DocFormer \cite{Appalaraju_2021_ICCV} instead proposed a discrete multimodal, focusing on sending visual and textual features through individual transformer layers. In each layer, visual and language features separately undergo self-attention with shared bounding box (bbox) information as a spatial feature.

Another multimodal architecture \cite{Dauphinee} classifies
using VGG16 and bag of words, respectively, for the visual and textual-based architecture. 
Their fusion technique has individual models running through their own architecture; later the class score vectors were concatenated. Finally, the resulting class score was classified by a meta-classifier using the XGBoost model.

RVL-CDIP \cite{rvl-cdip} is a benchmark dataset of letters, forms, emails, files, resumes, etc., with 400,000 grayscale images in 16 classes. It is split into training, validation, and test sets.
Dauphinee et al. \cite{Dauphinee} trained their model using this dataset and reported 93.03\% accuracy. Furthermore, DocFormer-base \cite{Appalaraju_2021_ICCV} achieved an accuracy of 96.17\%, which outperformed the 95.64\% accuracy of LayouLMv2 \cite{xu-etal-2021-layoutlmv2}.

Although these frameworks performed well on general document layout analysis, our experiments show that they do not work well for ETDs.
Hence, a framework based on fine-tuning YOLOv7 has been proposed to segment ETDs based on visual features. The framework was evaluated on a new dataset called ETD-OD \cite{ahuja-etal-2022-parsing} that consists of over 25,000 page images from 200 ETDs with manually drawn bboxes around objects (e.g., title, author, paragraph, etc.). However, lacking training samples led to low performance on minority categories such as date, algorithm, and equation.

To classify pages in legal documents, Wan et al. \cite{wan} proposed a text-based architecture, which used chunk embeddings (i.e., splitting the documents into multiple chunks), which were then used to train Doc2Vec to extract features. 
It achieved an overall accuracy of 97.97\%.

Our work attempts to overcome the limitations of existing frameworks by directly training a multimodal model with cross-attention using a dataset created by manual labeling and augmentation. We
employ a
hierarchical classification strategy to mitigate the sample imbalance problem.

\section{Methodology}
\subsection{Conceptual Overview}
In general, a multimodal workflow involves several unimodal neural networks to encode various input modalities independently. The extracted features are then combined using a fusion module. Finally, the fused features are fed into a classification network to make the prediction. There are three types of fusion \cite{baltrušaitis2017multimodal}: a) \textbf{Early Fusion} -- the features from each modality are combined at the start, and then the full model architecture is applied to the combined features; b) \textbf{Late Fusion} -- the individual modalities run through their own architecture, and the features from each modality are combined at the end to make the prediction; and c) \textbf{Hybrid Fusion} -- 
a combination of early and late fusion.
In this paper, we consider a two-stream multimodal model with a cross-attention layer by leveraging the early fusion technique. Figure~\ref{fig:framework} shows each modality extracting individual embeddings, with a projection layer to unify the dimensions. Then, leveraging the early fusion, we concatenated each projection and combined them with the features from the cross-attention layer. Finally, the full model was applied to the combined features.

\subsection{Visual Modality}
We used ResNetv2 \cite{resnetv2}, an improved version of ResNet \cite{resnet},
with 
the propagation formulation of the connections between the neural layers.
ResNETv2 has
new residual units with pre-activation.
Instead of putting batch normalization and ReLU after convolution, the authors put them prior to the convolution. ResNetv2 performed better than the original ResNet on the ImageNet-1k \cite{russakovsky2015imagenet} and CIFAR-10 / 100 datasets\footnote{\url{https://www.cs.toronto.edu/~kriz/cifar.html}}.

To use the ResNet50v2 model, we resized the ETD images to $224\times224$ pixels and fed them into the visual encoder. The output feature map was average-pooled to a fixed size with the width (W) and height (H). Later, it is flattened into a visual embedding sequence of length W$\times$H.

\subsection{Textual Modality}
For textual modality, we used BERT with the Talking-Heads Attention and the Gated GELU position-wise feed-forward networks \cite{bertwithTalkingHeads}, which is a transformer-based text encoder based on the original BERT \cite{bert} architecture by replacing the multi-head attention with  the talking-heads attention and replacing the ordinary dense layer with a gated linear unit with a GELU activation. The authors compared it against the Text-to-Text Transfer Transformer (T5) \cite{2020t5} model, which used multi-head attention. The results showed that talking-head-based models outperformed the multi-head attention on MNLI \cite{mnli} by at least 2\% on F1.

To generate the embeddings, we first use AWS Textract\footnote{\url{https://aws.amazon.com/textract/}} to extract text from the document images. We then used the pre-trained model of BERT with Talking-Heads Attention (large) \cite{bertwithTalkingHeads} from TensorFlow Hub as a text encoder. Using its pre-processing module, we performed tokenization and extracted the following features: a) \textbf{input type ids}, which identify which sequence a token belongs to when there is more than one sequence; b) \textbf{input mask}, which indicates whether a token should be attended to or not; and c) \textbf{input word ids}, which are the indices corresponding to each token in the sentence. These features are then fed through a trainable embedding layer.

\begin{table}[ht]
\caption{ETD500 -- page category labels (Category), total number of labeled pages (\#Pages), the number of augmented pages (\#Aug) (ToC = TableofContents, C. Abstract = ChapterAbstract).}
\centering
\begin{tabular}{c|c|c|l}
    \toprule
    {\bf Category} & {\bf \#Pages} & {\bf \#Aug} & {\bf Pages Description}\\
    \midrule
    Chapters & 71200 & - & content within sections labeled as chapters\\
    \midrule
    Appendices  & 9891 & - & detailed content not included in chapters\\
    \midrule
    References & 3385 & - & a list of biographical details of in-text citations\\
    \midrule
    ToC & 1114 & - & the list of chapters and page numbers\\
    \midrule
    TitlePage & 911 & - & a page containing a title and other metadata \\     
    \midrule
    Abstract & 777 & 602 & a narrative summary for the whole thesis\\
    \midrule
    ListofFigures & 586 & 651 & pages that include a ListofFigures and page numbers\\
    \midrule
    Ack & 543 & 542 & acknowledgment\\ 
    \midrule
    ListofTables & 477 & 584 & pages that include ListofTables and page numbers\\
    \midrule
    CV &  124 & 1116 & a curriculum vitae\\
    \midrule
    Dedication & 77 & 971 & devoting the work to motivating or supportive persons\\
    \midrule
    C. Abstract & 66 & 1518 & a chapter summary (occasional) \\
    \midrule
    Other & 3220 & - & pages that do not fit into any of the other 12 categories\\
    \midrule
    Total & 92,371 & 5,984 & - \\
    \midrule
    Subtotal (non-chapters) & 21,171 & - & - \\
    \bottomrule
\end{tabular}
\label{table1}
\end{table}

\subsection{Multimodal with Cross-Attention}
The vision and text encoders generate embeddings. However, the dimension of both embeddings needs to be unified for further early fusion. One general method that has been adopted to map the dimensions of these embeddings is applying a linear projection. Thus, we introduce a projection layer, where the embeddings from both modalities are combined. Our model takes one 256-D projection layer with 0.8 as a dropout rate \cite{dropout}. Later, the model fetches each embedding projection and performs an early fusion. To make our model focus on the most important pixels of an image that relate to their corresponding textual parts, we use the ``cross-attention'' \cite{Chen_2021_ICCV}. Cross-Attention \cite{Wei} combines asymmetrically two separate embedding sequences of two different modalities (i.e., visual and text). Further, we concatenate the early fusion of the projection layer with the attention sequence. We finally passed it through the ``softmax'' layer for the classification.

\begin{table*}[ht] 
    \caption{Performance on ETD samples in the test set -- a) performance of one-level classifier (i.e., \emph{Case a}), where ETDPC is trained on ETD500;
    b) performance of two-level classifier (i.e., \emph{Case b}), 
    training first on chapter vs. non-chapter pages, and next
    on the remaining categories, including 21,171 manually labeled samples; and c) performance of the two-level classifier (\emph{Case c}), trained on 21,171 manually labeled samples and 5,984 augmented samples. We highlight the categories with remarkable improvements of F1 scores.
    }
    \centering
    \begin{tabular}{c|c|c|c|c|c|c|c|c|c}
    \toprule
    \textbf{Category} & {\bf $\mathbf{P_{a}}$} & {\bf $\mathbf{R_{a}}$} & {\bf $\mathbf{F1_{a}}$} & {\bf $\mathbf{P_{b}}$} & {\bf $\mathbf{R_{b}}$} & {\bf $\mathbf{F1_{b}}$} & {\bf $\mathbf{P_{c}}$} & {\bf $\mathbf{R_{c}}$} & {\bf $\mathbf{F1_{c}}$}\\
     \midrule
    Chapters & 0.87 & 0.98 & {0.92} & -- & -- & -- & -- & -- & -- \\
    \midrule
    Appendices & 0.65 & 0.29 & \textcolor{red}{0.40} & 0.83 & 0.93 & {0.88} \textcolor{blue}{(+0.48)} & 0.83 & 0.93 & 0.88 \textcolor{teal}{(+0.00)} \\
    \midrule
    ReferenceList & 0.92 & 0.92 & {0.92} & 0.94 & 0.94 & {0.94} \textcolor{blue}{(+0.02)} & 0.95 & 0.94 & {0.95} \textcolor{teal}{(+0.01)} \\
    \midrule
    TableofContent & 0.81 & 0.75 & \textcolor{red}{0.78} & 0.80 & 0.86 & {0.83} \textcolor{blue}{(+0.05)} & 0.84 & 0.83 & {0.84} \textcolor{teal}{(+0.01)}\\
    \midrule
    TitlePage & 0.87 & 0.88 & \textcolor{red}{0.88} & 0.88 & 0.94 & {0.91} \textcolor{blue}{(+0.03)} & 0.87 & 0.94 & 0.91 \textcolor{teal}{(+0.00)} \\    
    \midrule
    Abstract & 0.33  & 0.02 & \textcolor{red}{0.03} & 0.60 & 0.50 & {0.54} \textcolor{blue}{(+0.51)} &  0.59 & 0.56 & {0.74} \textcolor{teal}{(+0.20)}\\
    \midrule
    ListofFigures & 0.60 & 0.57 & \textcolor{red}{0.58} & 0.66 & 0.64 & {0.65} \textcolor{blue}{(+0.07)} &  0.78 & 0.67 & {0.69} \textcolor{teal}{(+0.04)}\\
    \midrule
    Acknowledgment & 0.82 & 0.81 & {0.82} & 0.85 & 0.84 & {0.84} \textcolor{blue}{(+0.02)} & 0.88 & 0.84 & {0.93} \textcolor{teal}{(+0.09)}\\ 
    \midrule
    ListofTables & 0.58  & 0.35 & \textcolor{red}{0.44} & 0.65 & 0.44 & {0.52} \textcolor{blue}{(+0.08)} & 0.71 & 0.59 & {0.62} \textcolor{teal}{(+0.07)} \\
    \midrule
    CurriculumVitae & 0.83 & 0.26 & \textcolor{red}{0.40} & 0.92 & 0.58 & {0.71} \textcolor{blue}{(+0.31)} & 0.86 & 1.00 & {0.94} \textcolor{teal}{(+0.23)}\\
    \midrule
    Dedication & 1.00 & 0.27 & \textcolor{red}{0.43} & 1.00 & 0.55 & {0.71} \textcolor{blue}{(+0.28)} & 0.98 & 0.91 & {0.94} \textcolor{teal}{(+0.23)}\\
    \midrule
    ChapterAbstract & 0.00 & 0.00 & \textcolor{red}{0.00} & 0.00 & 0.00 & {0.00} \textcolor{blue}{(+0.00)} & 1.0 & 0.95 & {0.96} \textcolor{teal}{(+0.96)}\\
    \midrule
    Other & 0.61 & 0.20 & \textcolor{red}{0.30} & 0.75 & 0.55 & {0.63} \textcolor{blue}{(+0.33)} & 0.78 & 0.54 & {0.64} \textcolor{teal}{(+0.01)}\\
    \midrule
    macro F1 & 0.68 & 0.48 & \textcolor{red}{0.53} & 0.74 & 0.64 & 0.68 \textcolor{blue}{(+0.15)} & 0.83 & 0.81 & 0.83 \textcolor{teal}{(+0.15)} \\
    \bottomrule
    \end{tabular}
    \label{table2}
\end{table*}

\section{Experiments}\label{experiment}
\paragraph{Dataset:}
We compiled ETD500 \cite{choudhury2021automatic}, which consists of 500 scanned ETDs published between 1945 and 1990. There are 350 STEM and 150 non-STEM majors from 468 doctoral, 27 master’s, and 5 bachelor’s degrees. The dataset contains a total of 92,371 pages, available in PNG format. For the page-level classification task, we manually annotated all pages of ETD500 using an annotation tool developed by \cite{Caragea_Wu_Gollapalli_Giles_2016},
into 13 distinct category labels (Table~\ref{table1}). Later, OCR was performed on all pages using AWS Textract, a cloud-based service that detects and extracts texts from scanned documents. Textract converts PDF images into JSON containing text, ID, type (i.e., words or lines), bbox, and confidence score values.

\paragraph{Data Augmentation:}
Table~\ref{table1} gives the number of labeled pages for each category. The labeled data is highly skewed towards Chapters, Appendices, References, and Other pages, making it inappropriate to be used directly for training a machine learning model. Our goal was to perform data augmentation and increase the sample sizes to at least 1,000 for the minority categories (i.e., with fewer samples (Table~\ref{table1})) to mitigate the imbalance before training. We adopted the following strategies to generate pseudo-ETD training samples for the minority classes.
\begin{itemize}
    \item We first paraphrase the text extracted by the OCR.
    \item Second, we convert the text into images.
    \item Finally, we perform an image-based transformation.
\end{itemize}

For \textbf{paraphrasing} the ETD text, we adopted Google's \texttt{PEGASUS} \cite{PEGASUS}\footnote{\url{https://huggingface.co/docs/transformers/model_doc/pegasus}}, a pre-trained model for text summarization. The base architecture of PEGASUS \cite{PEGASUS} is a standard transformer encoder-decoder. In our experiment, we first use \texttt{PegasusTokenizer}, which is based on SentencePiece\footnote{\url{https://github.com/google/sentencepiece}}, to tokenize the input text while adding several parameters, including padding (i.e., pad to the longest sequence in the batch), max-length (i.e., pad to a maximum length specified with the argument max-length), and truncation (i.e.,  truncate to a maximum length specified with the argument max-length). To generate the paraphrase, we use the \texttt{PegasusForConditionalGeneration} model. After paraphrasing the text, we use a Python module called \texttt{textwrap}\footnote{\url https://docs.python.org/3/library/textwrap.html} to wrap the text with a width of 90, which represents the maximum length of wrapped lines. To \textbf{convert the text into images}, we 
change the fonts and size, adjust the textual position on a page, and finally draw
the text on an image using Python's \texttt{pillow} library\footnote{\url{https://pypi.org/project/Pillow/}}. To perform \textbf{image-based transformation}, we use a library called \texttt{ImgAug}\footnote{\url{https://github.com/aleju/imgaug}}. We adopted the following transformations from ScanBank \cite{scanbank} to generate the final pseudo images.

\begin{itemize}
    \item \textbf{Additive Gaussian noise} --  A flatbed scanner works by reflecting the light from paper and creating an image of the paper based on the naturally reflected light. Hence, we use Additive Gaussian Noise to mimic this effect. The parameters of this noise are heuristically chosen using trial and error.
    \item \textbf{Salt-and-pepper noise} -- Salt-and-pepper noise is often seen on images caused by sharp and sudden disturbances in the image signal. We heuristically chose 0.9 as the probability of replacing a pixel with noise.
    \item \textbf{Gaussian Blur} -- Unlike natural images, digital images must be encoded with a specified resolution resulting in a pre-determined number of bytes and some loss of sharpness. Therefore, we apply Gaussian blurring to smooth the images using a Gaussian Kernel, $\sigma{=0.5}$.
    \item \textbf{Linear Contrast} -- Although today’s scanners are built using modern technology, they cannot capture all colors of a natural object. To incorporate this scanning effect, we add Linear Contrast ($\alpha{=1}$).
\end{itemize}

\paragraph{Fine-tuning Hyper-parameters \& Training:}
We used our in-house high-performance computing cluster (HPC), which runs deep learning-based container service with various versions of TensorFlow and PyTorch.
We use a Tesla V100-SXM2 GPU to train our model and 12-core CPUs to perform other tasks, such as data augmentation. We heuristically fine-tuned the hyper-parameters. Our model with around 460M total parameters is trained on the original and augmented ETD pages. For the hyper-parameters, we choose the ``Adam'' optimizer with weight decay of 0.004, epsilon of 1e-07, clip value of 2.0, and learning rate of 0.001. In addition, we choose sparse categorical focal loss as the cost function. To avoid overfitting, we use the dropout rate of 0.8. We set up ``early stopping'' while monitoring the ``validation loss'' and the ``model checkpoint'' to save the best weight in each epoch. We finally trained the model with a batch size of 32 and 40 epochs, each taking around 1 hour.

\section{Evaluation and Results}
We split the manually labeled ETD pages into train, validation (25\%), and test sets (15\%) and consistently use the same test set for all evaluations.
\paragraph{Case a -- One-Level Classifier:}
The ETDPC model, applied to predict all of the 13 categories, achieved an overall accuracy of 84\% (Table~\ref{table2}). The 
classifier achieved low F1 scores on the categories with relatively small sample sizes 
than the categories with relatively large sample sizes.
Specifically, for ``Chapters'' it achieved an F1 score of 0.92.

\begin{figure*}[htp]
    \centering
    \subfloat[\centering]{{\includegraphics[width=12.8cm]{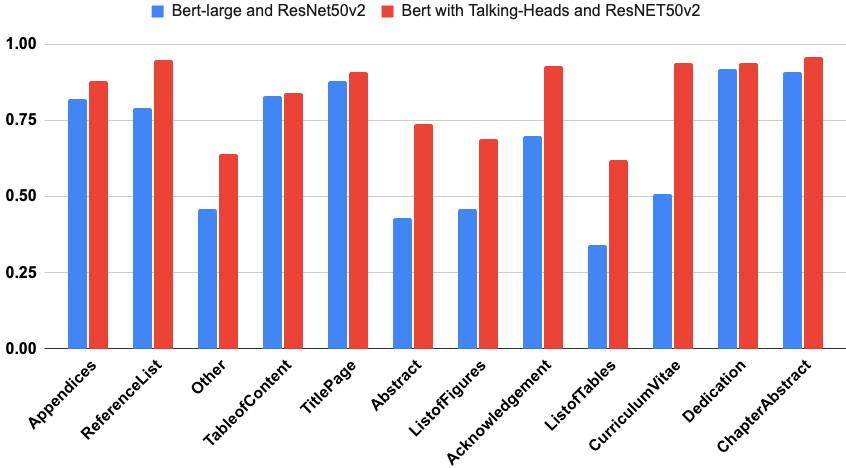} }}\\
    \subfloat[\centering]{{\includegraphics[width=12.8cm]{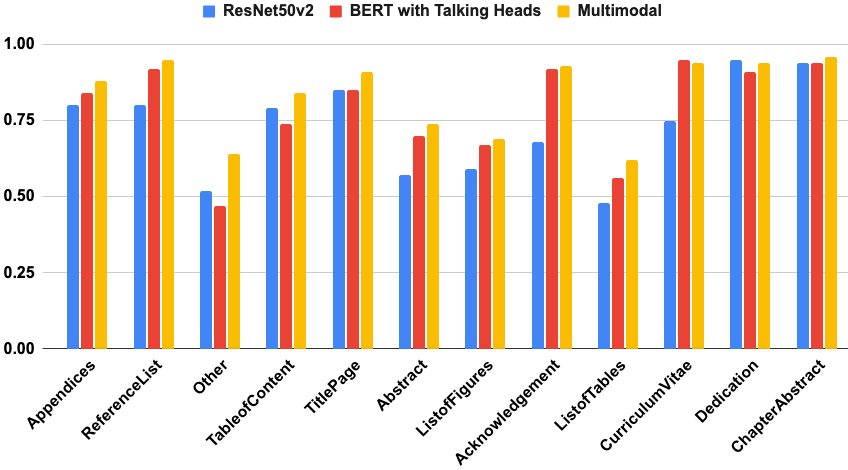}}}
    \caption{Ablation Study – a) Experiment 1 illustrates the performance increment when changing the original BERT model to BERT with Talking-Heads. b) Experiment 2 illustrates the performance of using the individual modalities vs. the multimodal model with cross-attention.}%
    \label{fig:ablation_study}%
\end{figure*}

\begin{figure}[ht]
    \centering
    \includegraphics[width=0.8\textwidth]{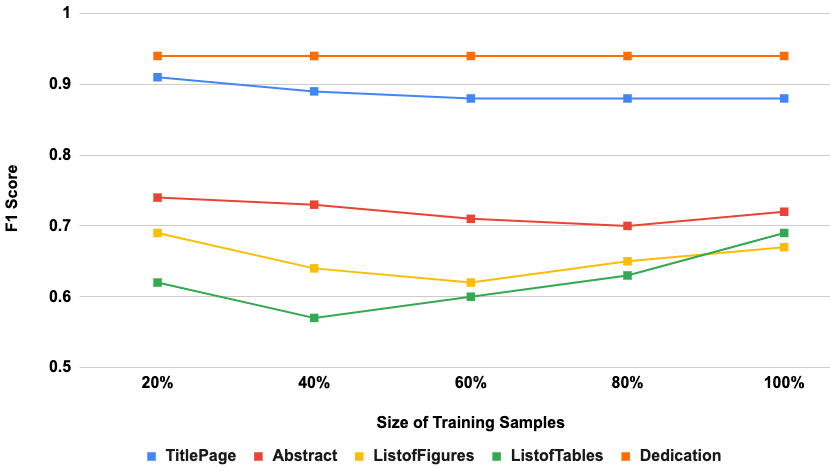}
    \caption{
    Diminishing returns in performance
    with increase in the number of training samples.}
    \label{fig:data_efficiency}
\end{figure}

\paragraph{Case b -- Two-Level Classifier:} To mitigate the data imbalance problems,
we train a two-level classifier. The first level classifies ETD pages into chapters and non-chapters. The second level classifier classifies non-chapter pages into Abstract, Acknowledgement, ListofFigures, ListofTables, Dedication, CV, and C. Abstract, etc., trained on 21,171 manually labeled pages. 

\paragraph{Case c -- Two-Level Classifier with Augmented Data: } To achieve better performance, we used the original training data, consisting of 21,171 manually labeled with 5,984 augmented training samples (Table~\ref{table1}).

\paragraph{Baseline Models:} For comparison, we fine-tuned DocFormer \cite{Appalaraju_2021_ICCV}, LayoutLMv2 \cite{xu-etal-2021-layoutlmv2},
and VGG16 \cite{vgg16} on the ETD500 dataset. LayoutLMv2 and DocFormer achieved low accuracy, below 30\% on the test dataset. Surprisingly, VGG16, a simple model based only on visual features, performed slightly better than LayoutLMv2 and DocFormer (Table~\ref{table3}).

\begin{table}[ht]
\caption{Comparison against baseline models. We report classification accuracy and macro F1 on the test set.}
\centering
\begin{threeparttable}
\begin{tabular}{c|c|c|c}
    \toprule
    {\bf Models} & {\bf \#param} & {\bf Accuracy} & {\bf macro F1} \\ 
    \midrule
    LayoutLMv2-base\tnote{1} & 200M & 0.09 & - \\
    \midrule
    DocFormer\tnote{2} & 174M & 0.28 & -\\
    \midrule
    VGG16 & 121M & 0.60 & 0.39\\
    \midrule
    ETDPC \textbf{(ours)} & 460M  & \textbf{0.85} & \textbf{0.83}\\
    \bottomrule
\end{tabular}
\begin{tablenotes}
\item[12] Due to the poor performance, only accuracy is reported.
\end{tablenotes}
\end{threeparttable}
\label{table3}
\end{table}

\paragraph{Results:}
Table~\ref{table3} shows that ETDPC significantly outperformed the baseline models, LayoutLMv2 and DocFormer, by 0.75 and 0.57 of accuracy, respectively. We also see a significant performance increase of 0.25 of the accuracy, and 0.44 of the F1 score compared to VGG16.

We report three types of performance in Table~\ref{table2}: a) performance of the one-level classifier, b) performance of the two-level classifier, and c) performance of the two-level classifier trained with augmented samples. The total inference time to predict the output took 13 minutes. Table~\ref{table2} indicates the one-level classifier performed well for classifying the ``Chapters'' category, achieving a 0.92 F1 score. The remaining categories achieve poor performance due to data imbalance. To mitigate data imbalance, we introduced a second-level classifier, which trained on 21,171 non-chapter manually labeled ETD pages.
The two-level classifier achieves improved performance, boosting the F1 score, ranging from 0.02 to 0.48 ($\mathbf{F1_{b}}$ in Table~\ref{table2}), depending on the categories. However, we still observed poor performance for the minority categories (i.e., containing less than 1K training samples (Table~\ref{table1})), including Abstract, ListofFigures, Acknowledgment, ListofTables, Curriculum Vitae, Dedication, and Chapter Abstract. Thus, we added augmented samples (Table~\ref{table1}) for these minority categories. Our result shows a significant increment in the performance, boosting the F1 scores up to 0.96, depending on the categories.

Table~\ref{table2} shows that ListofFigures and ListofTables achieved relatively low F1 scores. Upon further investigation, we observed 36 samples of ListofTables were misclassified as ListofFigures, whereas 23 samples of ListofFigures were misclassified as ListofTables. After randomly inspecting the testing samples, we found that they are visually very similar except for the page headings.
One way to improve the performance of these two categories is by integrating a heuristic-based method to capture lexical patterns, such as ``Table'' or ``Figure'', or the appearance of ``List of Figures'' or ``List of Tables'' in the headings. 

\subsection{Ablation Study}
We performed an ablation study of our proposed model. We conducted two ablation studies: Experiment 1: Changing the text encoder in the multimodal model; Experiment 2: Using individual modalities. We described each experiment in the following.

\paragraph{Experiment-1:} Figure~\ref{fig:ablation_study} (a) illustrates that when using the BERT with Talking-Heads Attention as a textual modality in our multi-modal model instead of BERT-large, ETDPC improved significantly, boosting the F1 score by 0.02 to 0.43, depending on the categories. 

\paragraph{Experiment-2:} For this experiment, we used the training samples in Case c. We used each individual modality for this experiment. Training with only visual modality took approximately 30 minutes. Training with only textual modality took approximately 14 hours. Training with our proposed multimodal with cross-attention mechanism took approximately 16 hours. Figure~\ref{fig:ablation_study} (b) shows that our proposed multimodal model with both modalities outperformed each individual modality, achieving 0.84 to 0.96 F1 scores for 9 out of 13 categories. The model achieves relatively low F1 scores for four categories , including ``ListofTables'', ``Other'', ``ListofFigures'', and ``Abstract'' with F1 scores ranging from 0.62 to 0.74. However, the performance of the multimodal model for these categories still outperformed each individual modality, improving F1 scores ranging from 0.10 -- 0.17 when compared to only visual modality and F1 scores ranging from 0.02 -- 0.17 when compared to only textual modality.

\subsection{Data Efficiency}
From Table~\ref{table2}, the F1 scores of several minority categories, such as ListofTables and ListofFigures, are below 0.70. We investigate whether the performance may improve when adding more training samples proportionally to these categories. Using the training dataset in Case c (i.e., 21,171 manually labeled and 5,984 augmented samples, a total of 27,155), we gradually increased the size of the training samples to the following minority categories: Abstract, Dedication, ListofFigures, ListofTables, and TitlePage, by 20\% each time.

We show the data efficiency for five categories in Figure~\ref{fig:data_efficiency}. The F1 score for the ListofTables category first decreased by $\sim0.05$ for the ListofTables and then increased by almost 0.10 when the training size was 100\%. 
The F1 score of the ListofFigures category reached the lowest when the training size is 60\% and then increased by about 0.05 when the training size was 100\%. 
The F1 score of the Abstract category decreased consistently until the training size is 80\% and then increased marginally by about $\sim0.02$.
The performance of the Dedication remained constant. 
The F1 score of the TitlePage category first decreased by $\sim0.02$, and then remained constant. 
Overall, the data efficiency analysis implies that the current datasizes are sufficient for most minority categories except for ListofTables and ListofFigures. In the future, we will add more training data and investigate various ways to data augmentation. 


\section{Conclusion \& Deployment Path}
We developed ETDPC, a framework aiming to classify ETD pages into 13 categories using a two-stream multimodal model with a cross-attention by leveraging vision (e.g., ResNet50v2) and language (e.g., BERT with Talking-Heads) models. The proposed model outperforms SOTA document page classification models. Although our model uses ETD500, consisting of scanned ETDs, for born-digital ETDs, it is straightforward to convert them to images. Our model takes 0.06 seconds to process a single ETD on average so it is scalable to millions of ETDs. Our model is also customizable. For example, we can easily adopt a multilingual language model to classify non-English ETDs. 

\paragraph{Deployment Path} 
The AI method we developed will be deployed following the steps below. First, we will integrate our model into a pipeline that extracts text using an open-source OCR package, such as docTR \cite{doctr2021}, that produces quality text and is more affordable than commercial OCR APIs, e.g., Amazon Textract. Second, we will deploy our framework on real-world data at the Old Dominion University Digital Commons hosting 3000+ ETDs and conduct an evaluation by manually inspecting the segmented ETDs from different years and academic disciplines. We will then mark and relabel incorrectly classified pages and fine-tune our pre-trained model using newly labeled pages. Finally, we will deploy the refined model on ETDs of the Old Dominion University and the Virginia Tech Libraries. Eventually, we will index ETD sections of different categories and make them available on the library websites. 




\section{Acknowledgments}
Support was provided by the Institute of Museum and Library Services through grant LG-37-19-0078-198. We also thank Himarsha R. Jayanetti for the data collection.

\bibliographystyle{unsrt}
\bibliography{references}

\begin{thebibliography}{10}

\bibitem{ahuja-etal-2022-parsing}
Aman Ahuja, Alan Devera, and Edward~Alan Fox.
\newblock Parsing electronic theses and dissertations using object detection.
\newblock In {\em Proceedings of the first Workshop on Information Extraction from Scientific Publications}, pages 121--130, Online, November 2022. Association for Computational Linguistics.

\bibitem{xu-etal-2021-layoutlmv2}
Yang Xu, Yiheng Xu, Tengchao Lv, Lei Cui, Furu Wei, Guoxin Wang, Yijuan Lu, Dinei Florencio, Cha Zhang, Wanxiang Che, Min Zhang, and Lidong Zhou.
\newblock {L}ayout{LM}v2: Multi-modal pre-training for visually-rich document understanding.
\newblock In {\em Proceedings of the 59th Annual Meeting of the Association for Computational Linguistics and the 11th International Joint Conference on Natural Language Processing (Volume 1: Long Papers)}, pages 2579--2591, Online, August 2021. Association for Computational Linguistics.

\bibitem{Appalaraju_2021_ICCV}
Srikar Appalaraju, Bhavan Jasani, Bhargava~Urala Kota, Yusheng Xie, and R.~Manmatha.
\newblock Docformer: End-to-end transformer for document understanding.
\newblock In {\em Proceedings of the IEEE/CVF International Conference on Computer Vision (ICCV)}, pages 993--1003, October 2021.

\bibitem{rvl-cdip}
Adam~W. Harley, Alex Ufkes, and Konstantinos~G. Derpanis.
\newblock Evaluation of deep convolutional nets for document image classification and retrieval.
\newblock In {\em 2015 13th International Conference on Document Analysis and Recognition (ICDAR)}, pages 991--995, 2015.

\bibitem{layoutlm}
Yiheng Xu, Minghao Li, Lei Cui, Shaohan Huang, Furu Wei, and Ming Zhou.
\newblock Layoutlm: Pre-training of text and layout for document image understanding.
\newblock In {\em Proceedings of the 26th ACM SIGKDD International Conference on Knowledge Discovery \& Data Mining}, KDD '20, page 1192–1200, New York, NY, USA, 2020. Association for Computing Machinery.

\bibitem{Dauphinee}
Tyler Dauphinee, Nikunj Patel, and Mohammad Rashidi.
\newblock Modular multimodal architecture for document classification.
\newblock {\em CoRR}, abs/1912.04376, 2019.

\bibitem{wan}
Lulu Wan, George Papageorgiou, Michael Seddon, and Mirko Bernardoni.
\newblock Long-length legal document classification.
\newblock {\em CoRR}, abs/1912.06905, 2019.

\bibitem{baltrušaitis2017multimodal}
Tadas Baltrušaitis, Chaitanya Ahuja, and Louis-Philippe Morency.
\newblock Multimodal machine learning: A survey and taxonomy, 2017.

\bibitem{resnetv2}
Kaiming He, Xiangyu Zhang, Shaoqing Ren, and Jian Sun.
\newblock Identity mappings in deep residual networks.
\newblock In Bastian Leibe, Jiri Matas, Nicu Sebe, and Max Welling, editors, {\em Computer Vision -- ECCV 2016}, pages 630--645, Cham, 2016. Springer International Publishing.

\bibitem{resnet}
Kaiming He, Xiangyu Zhang, Shaoqing Ren, and Jian Sun.
\newblock Deep residual learning for image recognition.
\newblock In {\em 2016 IEEE Conference on Computer Vision and Pattern Recognition (CVPR)}, pages 770--778, 2016.

\bibitem{russakovsky2015imagenet}
Olga Russakovsky, Jia Deng, Hao Su, Jonathan Krause, Sanjeev Satheesh, Sean Ma, Zhiheng Huang, Andrej Karpathy, Aditya Khosla, Michael Bernstein, Alexander~C. Berg, and Li~Fei-Fei.
\newblock Imagenet large scale visual recognition challenge, 2015.

\bibitem{bertwithTalkingHeads}
Noam Shazeer, Zhenzhong Lan, Youlong Cheng, Nan Ding, and Le~Hou.
\newblock Talking-heads attention, 2020.

\bibitem{bert}
Jacob Devlin, Ming{-}Wei Chang, Kenton Lee, and Kristina Toutanova.
\newblock {BERT:} pre-training of deep bidirectional transformers for language understanding.
\newblock In Jill Burstein, Christy Doran, and Thamar Solorio, editors, {\em Proceedings of the 2019 Conference of the North American Chapter of the Association for Computational Linguistics: Human Language Technologies, {NAACL-HLT} 2019, Minneapolis, MN, USA, June 2-7, 2019, Volume 1 (Long and Short Papers)}, pages 4171--4186. Association for Computational Linguistics, 2019.

\bibitem{2020t5}
Colin Raffel, Noam Shazeer, Adam Roberts, Katherine Lee, Sharan Narang, Michael Matena, Yanqi Zhou, Wei Li, and Peter~J. Liu.
\newblock Exploring the limits of transfer learning with a unified text-to-text transformer.
\newblock {\em Journal of Machine Learning Research}, 21(140):1--67, 2020.

\bibitem{mnli}
Seonhoon Kim, Inho Kang, and Nojun Kwak.
\newblock Semantic sentence matching with densely-connected recurrent and co-attentive information, 2018.

\bibitem{dropout}
Nitish Srivastava, Geoffrey Hinton, Alex Krizhevsky, Ilya Sutskever, and Ruslan Salakhutdinov.
\newblock Dropout: A simple way to prevent neural networks from overfitting.
\newblock {\em Journal of Machine Learning Research}, 15(56):1929--1958, 2014.

\bibitem{Chen_2021_ICCV}
Chun-Fu~(Richard) Chen, Quanfu Fan, and Rameswar Panda.
\newblock Crossvit: Cross-attention multi-scale vision transformer for image classification.
\newblock In {\em Proceedings of the IEEE/CVF International Conference on Computer Vision (ICCV)}, pages 357--366, October 2021.

\bibitem{Wei}
Xi~Wei, Tianzhu Zhang, Yan Li, Yongdong Zhang, and Feng Wu.
\newblock Multi-modality cross attention network for image and sentence matching.
\newblock In {\em 2020 IEEE/CVF Conference on Computer Vision and Pattern Recognition (CVPR)}, pages 10938--10947, 2020.

\bibitem{choudhury2021automatic}
Muntabir Hasan~Choudhury, Himarsha~R. Jayanetti, Jian Wu, William~A. Ingram, and Edward~A. Fox.
\newblock Automatic metadata extraction incorporating visual features from scanned electronic theses and dissertations.
\newblock In {\em 2021 ACM/IEEE Joint Conference on Digital Libraries (JCDL)}, pages 230--233, 2021.

\bibitem{Caragea_Wu_Gollapalli_Giles_2016}
Cornelia Caragea, Jian Wu, Sujatha Gollapalli, and C.~Giles.
\newblock Document type classification in online digital libraries.
\newblock {\em Proceedings of the AAAI Conference on Artificial Intelligence}, 30(2):3997--4002, Feb. 2016.

\bibitem{PEGASUS}
Jingqing Zhang, Yao Zhao, Mohammad Saleh, and Peter~J. Liu.
\newblock Pegasus: Pre-training with extracted gap-sentences for abstractive summarization.
\newblock In {\em Proceedings of the 37th International Conference on Machine Learning}, ICML'20. JMLR.org, 2020.

\bibitem{scanbank}
S.~Kahu, W.~A. Ingram, E.~A. Fox, and J.~Wu.
\newblock Scanbank: A benchmark dataset for figure extraction from scanned electronic theses and dissertations.
\newblock In {\em 2021 ACM/IEEE Joint Conference on Digital Libraries (JCDL)}, pages 180--191, Los Alamitos, CA, USA, sep 2021. IEEE Computer Society.

\bibitem{vgg16}
Karen Simonyan and Andrew Zisserman.
\newblock Very deep convolutional networks for large-scale image recognition, 2015.

\bibitem{doctr2021}
Mindee.
\newblock doctr: Document text recognition.
\newblock \url{https://github.com/mindee/doctr}, 2021.

\end{thebibliography}
\end{document}